\documentclass[letterpaper, 10 pt, conference]{ieeeconf}  

\IEEEoverridecommandlockouts                              

\usepackage{cite}
\usepackage{amsmath,amssymb,amsfonts}
\usepackage{algorithmic}
\usepackage{graphicx}
\usepackage{textcomp}
\usepackage{xcolor}
\usepackage{multirow}
\usepackage{url}

\def\ie{\emph{i.e}.}

\title{\LARGE \bf
Neural Radiance Fields for Novel View Synthesis \\ in Monocular Gastroscopy
}

\author{Zijie Jiang$^{1}$, Yusuke Monno$^{1}$, Masatoshi Okutomi$^{1}$, Sho Suzuki$^{2}$, and  Kenji Miki$^{3}$
\thanks{$^{1}$Z. Jiang, Y. Monno, M. Okutomi are with the Department of Systems and Control Engineering, School of Engineering, Tokyo Institute of Technology, Meguro-ku, Tokyo 152-8550, Japan (email: { \{zjiang, ymonno\}@ok.sc.e.titech.ac.jp; mxo@sc.e.titech.ac.jp}).}
\thanks{$^{2}$S. Suzuki is with the Department of Gastroenterology, International University of Health and Welfare Ichikawa Hospital, Ichikawa-shi, Chiba, 272-0827, Japan.}
\thanks{$^{3}$K. Miki is with the Department of Internal Medicine, Tsujinaka Hospital Kashiwanoha, Kashiwa-city, Chiba 277-0871, Japan.}
}

\begin{document}

\maketitle
\thispagestyle{empty}
\pagestyle{empty}

\begin{abstract}
Enabling the synthesis of arbitrarily novel viewpoint images within a patient's stomach from pre-captured monocular gastroscopic images is a promising topic in stomach diagnosis.
Typical methods to achieve this objective integrate traditional 3D reconstruction techniques, including structure-from-motion (SfM) and Poisson surface reconstruction.
These methods produce explicit 3D representations, such as point clouds and meshes, thereby enabling the rendering of the images from novel viewpoints.
However, the existence of low-texture and non-Lambertian regions within the stomach often results in noisy and incomplete reconstructions of point clouds and meshes, hindering the attainment of high-quality image rendering.
In this paper, we apply the emerging technique of neural radiance fields (NeRF) to monocular gastroscopic data for synthesizing photo-realistic images for novel viewpoints.
To address the performance degradation due to view sparsity in local regions of monocular gastroscopy, we incorporate geometry priors from a pre-reconstructed point cloud into the training of NeRF, which introduces a novel geometry-based loss to both pre-captured observed views and generated unobserved views.
Compared to other recent NeRF methods, our approach showcases high-fidelity image renderings from novel viewpoints within the stomach both qualitatively and quantitatively.
\end{abstract}

\section{INTRODUCTION}

\begin{figure*}[htbp]
    \centering
    \includegraphics[width=0.98\linewidth]{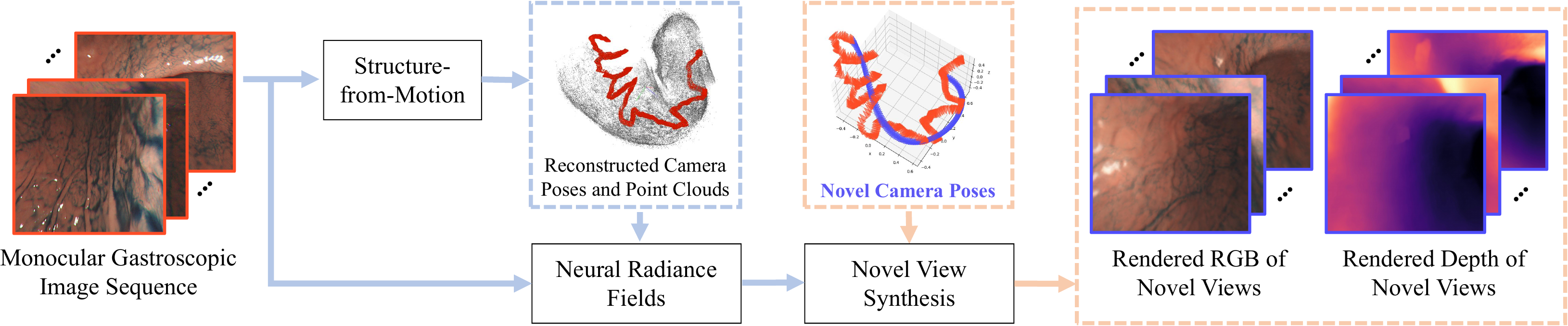}
    \caption{\textbf{The overall process flow.} Using a real monocular gastroscopic image sequence, we first apply structure-from-motion (SfM) to obtain camera poses and a reconstructed point cloud. Then, we train neural radiance fields (NeRF) of the stomach, where we propose a novel geometry-based loss exploiting the point cloud from SfM. In the application phase, RGB images and depth maps of novel views can be synthesized through volume rendering of NeRF.
    }
    \label{fig:pipeline}
\end{figure*}

\begin{figure}[htbp]
    \centering
    \includegraphics[width=1\linewidth]{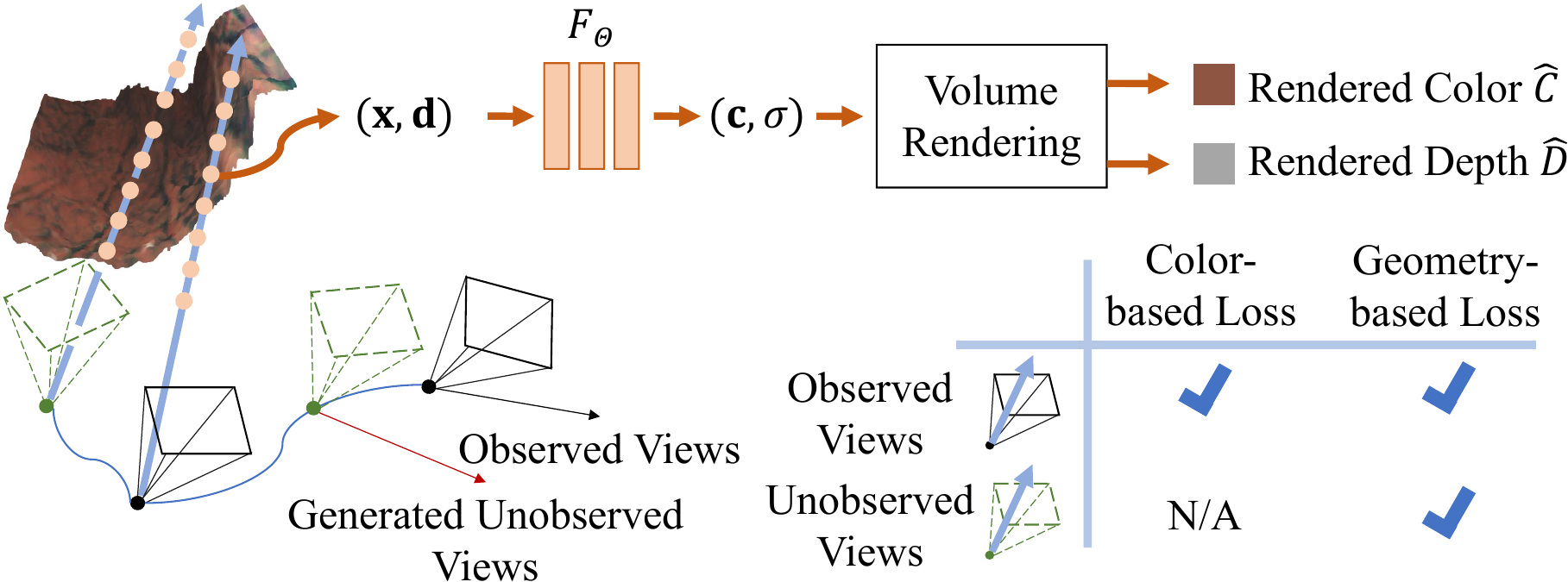}
    \caption{\textbf{The overview of our proposed NeRF method.}
    As a standard NeRF method, we train a network $F_{\Theta}$ to estimate the color $\bf{c}$ and the density $\sigma$ given the 3D point coordinate $\bf{x}$ and the viewing direction $\bf{d}$ as inputs. The key of our method is twofold: 1)~We generate unobserved views by interpolating consecutive observed views to address view sparsity and 2)~we apply a geometry-based loss for both observed and unobserved views to effectively constrain the learned geometry by using the point cloud reconstructed by SfM. The technical details are in the methodology section.
    }
    \label{fig:training}
\end{figure}

Gastroscopy plays a crucial role in minimally invasive diagnostic applications.
It captures rich 2D RGB information within the patient's gastric cavity, which offers the practitioners valuable assistance in clinical diagnosis and intervention for various pathological conditions.
However, a notable limitation in gastroscopic examinations lies in the constrained viewpoints and viewing angles that the practitioners have inside the stomach, which are determined by the trajectory of the gastric endoscope.
This typically impedes the practitioners from obtaining adjustable and comprehensive observations within the gastric cavity.

A typical solution to enable generating free-viewpoint observations, \ie, novel view synthesis, within the stomach involves reconstructing a 3D model of the stomach based on the captured gastroscopic images.
Structure-from-motion (SfM)~\cite{snavely2006photo, pollefeys2008detailed, schonberger2016structure} is a general technique used to recover camera poses and generate a 3D point cloud of the captured scene from image collections, which has been applied in several studies to reconstruct the 3D model of a target organ from an endoscope video~\cite{mills2014hierarchical, sun2013surface, lurie20173d, furukawa2016shape, alcantarilla2013enhanced, widya20193d}.
The method in~\cite{widya20193d} successfully reconstructed camera poses and the 3D model of an entire stomach from a standard monocular gastroscopic video by investigating the combined effect of chromo-endoscopy and color channel selection on SfM.
For the enhanced visualization of the reconstructed 3D model, the authors further utilized Poisson surface reconstruction~\cite{kazhdan2013screened} to generate textured meshes from the 3D point cloud acquired through SfM.
Although novel view synthesis can be achieved from their reconstructed textured 3D meshes, the reconstructed 3D model often exhibits noise and incompleteness due to the existence of low-texture and non-Lambertian regions in gastroscopic images, which consequently leads to low-quality image synthesis.

Recently, neural radiance fields (NeRF)~\cite{mildenhall2020nerf, barron2023zipnerf} have shown significant progress in the tasks of novel view synthesis and 3D reconstruction from the images with known camera poses.
In contrast to traditional 3D reconstruction methods producing explicit and discrete 3D representations such as point clouds and meshes, NeRF learns an implicit and continuous representation of the scene appearance and geometry, which is encoded in the parameters of multi-layer perceptrons (MLPs).
The MLP network takes a 3D point position and a 2D camera viewing direction as inputs, predicting the corresponding RGB color and density information.
The observation of the 3D scene from an arbitrary viewpoint can finally be obtained through the integration of color and density information along cast camera rays using volume rendering~\cite{kajiya1984ray}.
The emerging NeRF technique has been applied to diverse medical domains, such as 3D reconstruction of deformable tissues from single-viewpoint stereo endoscopy~\cite{wang2022neural, zha2023endosurf} or monocular endoscopy~\cite{batlle2023lightneus}, computed tomography~\cite{reed2021dynamic, corona2022mednerf}, and magnetic resonance imaging~\cite{wu2022arbitrary, shen2022nerp}.
However, this technique is still not fully explored and evaluated in the context of novel view synthesis based on monocular gastroscopy.

In this paper, we explore the application of NeRF to monocular gastroscopic data to achieve high-quality results in novel view synthesis.
We observe that directly applying a state-of-the-art NeRF method~\cite{barron2023zipnerf} trained with standard color-based losses to monocular gastroscopic data results in broken geometry and blurry image rendering.
One main reason for the performance degradation is attributed to the view sparsity in local regions present in gastroscopic data, which results in the insufficiency of the color-based loss to address the shape-radiance ambiguity during the training of NeRF.
To enhance the training results on monocular gastroscopic data, we incorporate the color-based loss with an additional geometry-based loss, which exploits the supervision using a point cloud pre-reconstructed by SfM, sharing a similar idea with DS-NeRF~\cite{deng2022depth}.
Notably, our proposed geometry-based supervision differs from the one in DS-NeRF mainly in two aspects:
1) In addition to utilizing the sparse depth maps obtained from a point cloud for supervision, we also incorporate a depth smoothness loss based on the shape priors of the stomach.
2) DS-NeRF imposes depth supervision solely to pre-captured observed views, whereas our method imposes depth supervision to both pre-captured observed views and unobserved views randomly interpolated from observed views, which better regularizes the learned geometry of NeRF.
Our experimental results showcase the effectiveness of our geometry-based supervision in enhancing both rendering quality and recovered geometry compared to existing methods.

\section{Methodology}

The overall flow of our method is illustrated in Fig.~\ref{fig:pipeline}.
Given a real monocular gastroscopic image sequence, our method initially follows the SfM steps proposed in~\cite{widya20193d} to reconstruct the camera poses and the 3D point cloud of the stomach. Then, NeRF is trained using the image sequence, the camera poses, and the point cloud.
In the application phase, we render RGB images and depth maps for novel camera poses using the learned NeRF of the stomach.

Figure~\ref{fig:training} illustrates the overview of our proposed NeRF method.
We train a network $F_{\Theta}$ which takes the 3D point coordinate $\bf{x}$ and the viewing direction $\bf{d}$ as inputs and outputs the color $\bf{c}$ and the density $\sigma$ at this 3D coordinate $\bf{x}$.
The key of our method is summarized as follows: 1)~We generate unobserved views by interpolating consecutive observed views to address view sparsity and 2)~we apply a geometry-based loss for both observed and unobserved views to effectively constrain the learned geometry by using the point cloud pre-reconstructed by SfM.
The technical details are explained in the following subsections.

\subsection{Unobserved View Generation}
\label{subsec:2-1}

After obtaining the camera poses of all observed views through SfM, we generate $k$ unobserved views between each consecutive observed view pair.
We parameterize the camera pose $P$ by $P=(\mathbf{t}, q)$, where $\mathbf{t}\in\mathbb{R}^3$ is the 3D position and $q$ is the unit quaternion representing rotation.
Given a pair of consecutive observed views with camera poses $P_b$ and $P_{b+1}$, we interpolate the 3D position and the rotation separately to generate a new camera pose $P_{k}=(\mathbf{t}_{k},q_{k})$ of an unobserved view as
\begin{equation}
\begin{split}
    \mathbf{t}_{k}=(1-\alpha_k)\mathbf{t}_{b}+\alpha_k \mathbf{t}_{b+1}, \\
    q_{k}=q_b(q_{b}^{-1}q_{b+1})^{\alpha_k},
\end{split}
\end{equation}
where $\alpha_k$ is randomly sampled from 0 to 1, which determines the ratio between the distances from the generated camera pose $P_k$ to $P_b$ and $P_k$ to $P_{b+1}$.
To maximize the utilization of additional geometric constraints brought by unobserved views, we regenerate new unobserved views and replace the previously used unobserved views with them every 2,000 training iterations in our experiments.

\subsection{Ray Sampling and Volume Rendering}
\label{subsec:2-2}

In each iteration of NeRF training, we sample two types of camera rays, denoted as the color ray and the depth ray.
The color ray passes through the camera origin and a pixel on the image plane, with its ground-truth color value being the color of the corresponding pixel of the real image.
The depth ray passes through the camera origin and a reconstructed 3D point visible from the camera, with its reference depth value being the transformed depth of the corresponding 3D point in the camera coordinate.
Since the ground-truth color is only available for the observed views, we sample the set of color rays $\mathcal{R}_c^{ob}$ only on the observed views.
The set of depth rays are separately sampled on both the observed views and unobserved views, denoted as $\mathcal{R}_d^{ob}$ and $\mathcal{R}_d^{nv}$ respectively.

Given a camera ray $\mathbf{r}(t)=\mathbf{o}+t\mathbf{d}$ from the camera origin $\mathbf{o}$ with the viewing direction $\mathbf{d}$, we compute the color of this camera ray using volume rendering as described in~\cite{mildenhall2020nerf}:
\begin{align}
  \hat{C}(\mathbf{r})=\int_{t_n}^{t_f}T(t)\sigma(\mathbf{r}(t))\mathbf{c}(\mathbf{r}(t),\mathbf{d})\mathrm{d}t,
\end{align}
where $T(t)=\mathrm{exp}(-\int_{t_n}^{t}\sigma(\mathbf{r}(s))\mathrm{d}s$,
$t_n$ and $t_f$ are the near and far bounds of the traveled distance $t$.
The density $\sigma$ and color $\mathbf{c}$ are predicted by the MLP network $F_{\Theta}: (\mathbf{r}(t), \mathbf{d})\rightarrow (\sigma, \mathbf{c})$.
Similarly, the depth of this camera ray can be rendered as
\begin{align}
  \hat{D}(\mathbf{r})=\int_{t_n}^{t_f}T(t)\sigma(\mathbf{r}(t))t\mathrm{d}t.
\end{align}
For the computing of training loss, we render both the color and depth for camera rays in $\mathcal{R}_c^{ob}$ and only render the depth for camera rays in $\mathcal{R}_d^{ob} \cup \mathcal{R}_d^{nv}$.

\subsection{Training Loss}
\label{subsec:2-3}
Given sampled sets of camera rays $\mathcal{R}_c^{ob}$, $\mathcal{R}_d^{ob}$ and $\mathcal{R}_d^{nv}$, and their rendered results, the training loss of our method is computed as follows.

\noindent
\textbf{Color-based loss.} The color-based loss is computed from camera rays in $\mathcal{R}_c^{ob}$ and is formulated as:
\begin{align}
  L_{color}^{ob}=\sum_{\mathbf{r}\in\mathcal{R}_{c}^{ob}}\Vert\hat{C}(\mathbf{r})-C(\mathbf{r})\Vert_2^2,
\end{align}
where $\hat{C}(\mathbf{r})$ is the rendered color of the camera ray in $\mathcal{R}_{c}^{ob}$ and $C(\mathbf{r})$ is the corresponding ground-truth color of a real training image.

\noindent
\textbf{Geometry-based loss for observed views.}
For each camera ray $\mathbf{r}$ in $\mathcal{R}_d^{ob}$, we first compute the difference between the rendered depth $\hat{D}(\mathbf{r})$ and its reference depth value $D_{pc}(\mathbf{r})$ obtained from the reconstructed 3D point as
\begin{align}
    l_d(\mathbf{r})=\Vert\hat{D}(\mathbf{r})-D_{pc}(\mathbf{r})\Vert_2^2.
\end{align}
We further introduce the smoothness loss to enforce local smoothness in the rendered depth of camera rays either in $\mathcal{R}_c^{ob}$ or $\mathcal{R}_d^{ob}$, considering that the internal structure of the stomach is typically smooth and continuous:
\begin{align}
    l_s(\mathbf{r})=\vert \partial_u\hat{D}(\mathbf{r}) \vert + \vert \partial_v\hat{D}(\mathbf{r}) \vert,
\end{align}
where $\partial_u$ represents the derivative operation in the horizontal direction on the image plane and $\partial_v$ represents the derivative operation in the vertical direction on the image plane.
We also combine the KL divergence loss $l_{KL}(\mathbf{r})$ introduced in~\cite{deng2022depth} to constrain the ray distribution to be unimodal. Please refer to~\cite{deng2022depth} for further details on $l_{KL}(\mathbf{r})$.
Our final geometry-based loss for the observed views is formulated as:
\begin{align}
  \begin{split}
    L_{depth}^{ob}=\sum_{\mathbf{r}\in\mathcal{R}_{d}^{ob}}&\lambda_{d}\Vert\hat{D}(\mathbf{r})-D_{pc}(\mathbf{r})\Vert_2^2 + \lambda_{KL}l_{KL}(\mathbf{r}) \\
    &+ \sum_{\mathbf{r}\in\mathcal{R}_{c}^{ob} \cup \mathcal{R}_{d}^{ob}}\lambda_{s}l_s(\mathbf{r}),
  \end{split}
\end{align}
where $\lambda_d$, $\lambda_{KL}$ and $\lambda_{s}$ are hyperparameters.

\noindent
\textbf{Geometry-based loss for unobserved views.}
Applying geometry-based loss solely on the observed views may sometimes be ineffective in constraining the learned 3D geometry of NeRF due to the sparse depth supervision signals obtained from the reconstructed 3D point cloud.
Thus, we propose to integrate additional geometry-based supervision on generated unobserved views to further regularize the learned geometry, which is formulated as
\begin{align}
  \begin{split}
    L_{depth}^{nv}=\sum_{\mathbf{r}\in\mathcal{R}_{d}^{ob}}&\lambda_{d}\Vert\hat{D}(\mathbf{r})-D_{pc}(\mathbf{r})\Vert_2^2 \\
    &+ \lambda_{KL}l_{KL}(\mathbf{r}) + \lambda_{s}I_s(\mathbf{r}).
  \end{split}
\end{align}

The final training loss of our method is concluded as
\begin{align}
L_{total}=L_{color}^{ob}+L_{depth}^{ob}+L_{depth}^{nv}.
\end{align}

\vspace{1mm}
\section{Experimental Results}

\subsection{Datasets and Implementation Details}
We evaluated our method using two monocular gastroscopic videos within two different subjects, identified as Seq. A and Seq. B, obtained in our previous study~\cite{widya20193d}. The data acquisition protocol in~\cite{widya20193d} and the experimental protocol in this study were approved by the ethics committees of the related institutions. The informed consent was obtained from all participating subjects. Seq. A and B consist of a total of 833 images and 424 images, respectively.

We implemented our method based on Zip-NeRF~\cite{barron2023zipnerf} by incorporating our proposed geometry-based loss for observed and unobserved views.
During the training, we generated $k=2$ unobserved views between each consecutive observed view pair.
We sampled 8,192 camera rays for $\mathcal{R}^{ob}_c$ and 4,096 depth rays for $\mathcal{R}^{ob}_d$ and $\mathcal{R}_d^{nv}$ separately.
The hyperparameters in training loss were experimentally set as $\lambda_d=10$, $\lambda_{KL}=0.1$, and $\lambda_{s}=10$.
The number of training iterations was set to 25,000 and the other hyperparameters were set the same as Zip-NeRF.
All experiments were performed on a single Nvidia RTX-3090 GPU.

\subsection{Rendering Results for A Novel Camera Trajectory}
\label{sec:3-1}

\begin{figure}[t!]
    \centering
    \includegraphics[width=1\linewidth]{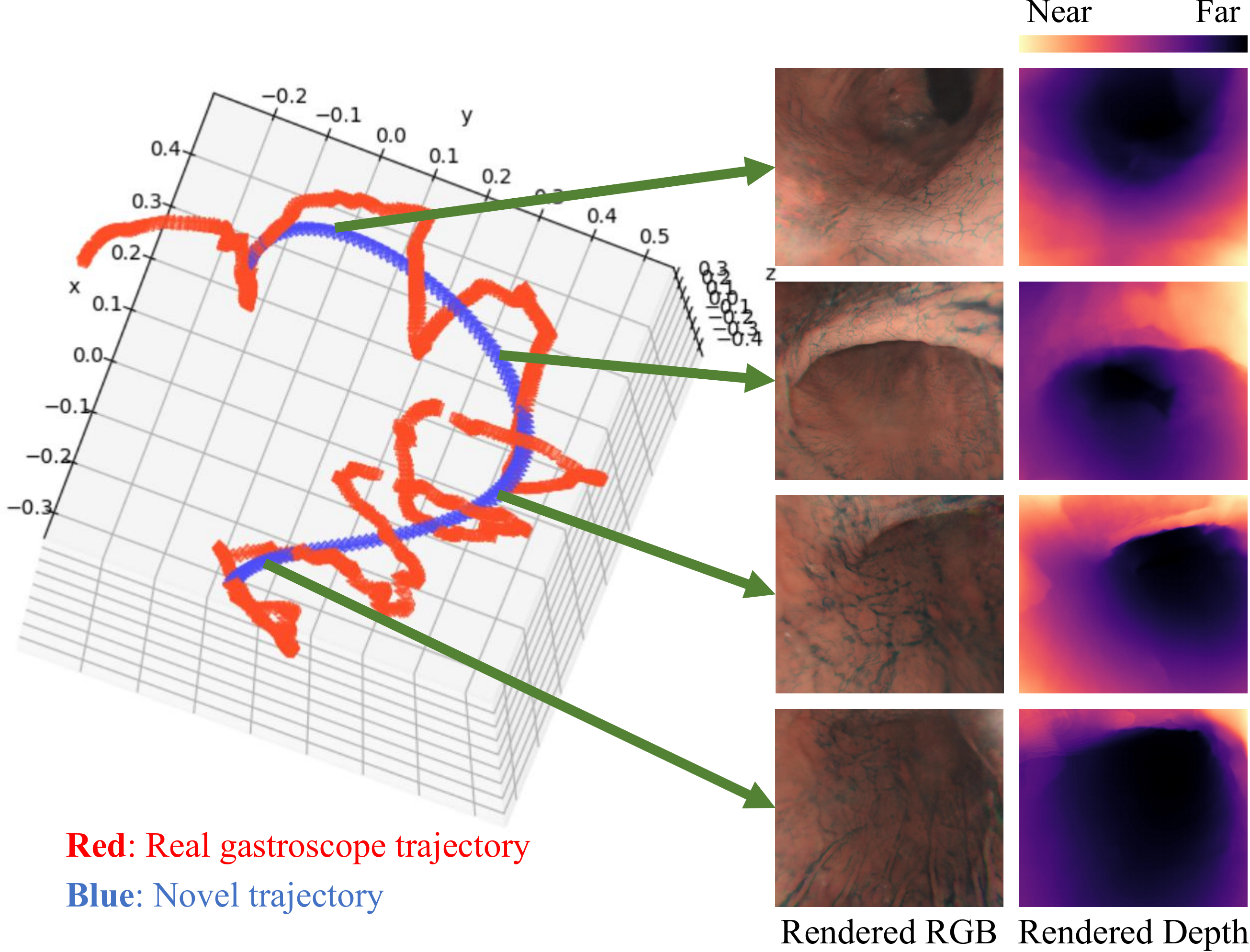}
    \caption{\textbf{Rendering results for a novel camera trajectory.} The camera trajectory in \textcolor{red}{red} color represents a real gastroscope trajectory, which was used for training NeRF. The camera trajectory in \textcolor{blue}{blue} color represents a novel trajectory for the view synthesis application.}
    \label{fig_novel_traj}
\end{figure}

We first trained NeRF using all images in each sequence.
Given the learned NeRF of the stomach, we can render RGB images and depth maps within the stomach from viewpoints entirely different from those in the real gastroscopy image sequence, as illustrated in Fig.~\ref{fig_novel_traj}.
We observe that even when the novel viewpoint is at a considerable distance from the camera poses of the real gastroscopy image sequence, our method can obtain clear RGB images and plausible depth maps.
This capability enables the practitioners to freely adjust the viewing trajectory to obtain the best observations inside the stomach.

\subsection{Comparison of Rendered RGB Images}
\label{sec:3-2}

For numerical and visual comparisons using ground-truth images, we conducted another experiment by recomposing the original image sequence into training images and testing images. Specifically, we first reduce the frame rate to one-fourth to evaluate the effectiveness of our method for view sparsity. Then, we selected every second frame for the testing and used the remaining frames for the training. As a consequence, for Seq. A, two sets of 104 images were used for the training and the testing, respectively. Similarly, for Seq. B, two sets of 53 images were used for the training and the testing, respectively.

\begin{figure}[!t]
    \centering
    \includegraphics[width=1\linewidth]{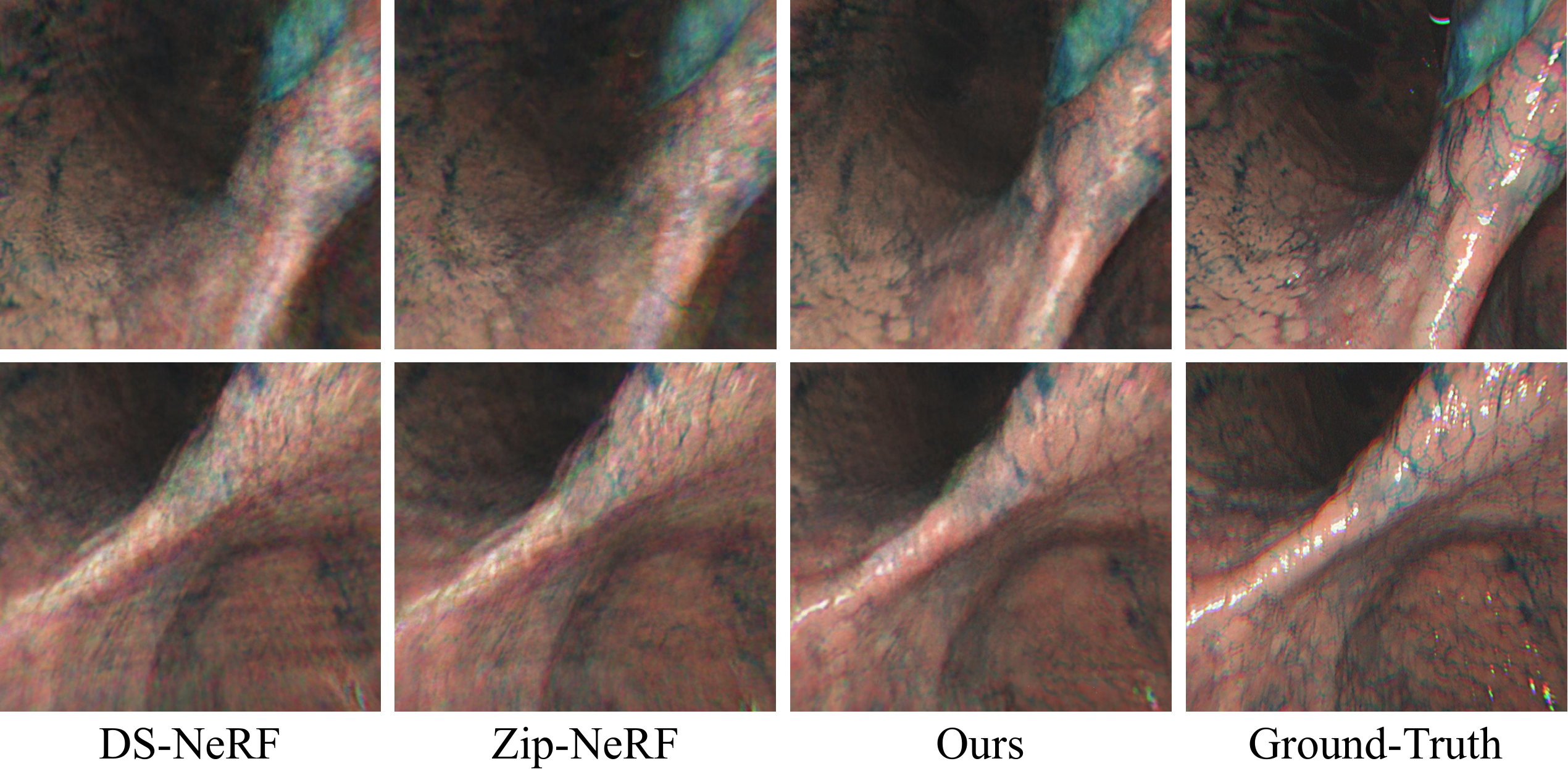}
    \caption{\textbf{The qualitative comparisons of rendered RGB images.} The top and the second rows show the results for two different viewpoints in the testing images.}
    \label{fig_2}
\end{figure}

We compared our method with two most related NeRF methods: Zip-NeRF~\cite{barron2023zipnerf} and DS-NeRF~\cite{deng2022depth}.
Zip-NeRF is our baseline, which only uses color-based loss for training.
DS-NeRF is based on the original NeRF~\cite{mildenhall2020nerf} and leverages the reconstructed point clouds to constrain the learned geometry similarly, but without considering unobserved views and smoothness loss.
Table~\ref{tab_1} presents quantitative comparisons of the rendered RGB image quality, where we evaluate the average peak-to-signal raito~(PSNR) and structural similarity index measure (SSIM) for all the testing images.
Across both sequences, our method attains consistently superior results in synthesizing high-quality RGB images for novel viewpoints not included in the training.
The improvement from the baseline Zip-NeRF shows the positive impact of the proposed geometry-based loss for monocular gastroscopic data.
Figure~\ref{fig_2} provides qualitative comparisons of the rendered RGB images for two testing views.
Our method produces more photorealistically synthesized images for novel viewpoints compared to the other methods, including more image details such as sharp edges between foreground and background.

\subsection{Comparison of Rendered Depth Mpas}
\label{sec:3-3}

\begin{table}[t!]
\caption{\textbf{The quantitative evaluation of rendered RGB images.} The best results are highlighted using \textbf{bold} formatting.}
\label{tab_1}
\centering
\renewcommand\arraystretch{1.2}
\resizebox{\linewidth}{!}{
\begin{tabular}{l|cc|cc}
\hline
\multirow{2}{*}{}         & \multicolumn{2}{c|}{Seq. A} & \multicolumn{2}{c}{Seq. B} \\ \cline{2-5} 
                          & PSNR$\uparrow$         & SSIM$\uparrow$          & PSNR$\uparrow$          & SSIM$\uparrow$         \\ \hline
DS-NeRF~\cite{deng2022depth}                   & 22.49          & 0.838          & 20.35          & 0.694         \\
Zip-NeRF~\cite{barron2023zipnerf}                  & 25.07        & 0.856        & 22.78        & 0.751       \\
Ours                      & \textbf{26.73}        & \textbf{0.870}        & \textbf{23.37}        & \textbf{0.767}       \\ \hline
\end{tabular}
}
\end{table}

\begin{figure}[t!]
    \centering
    \includegraphics[width=1\linewidth]{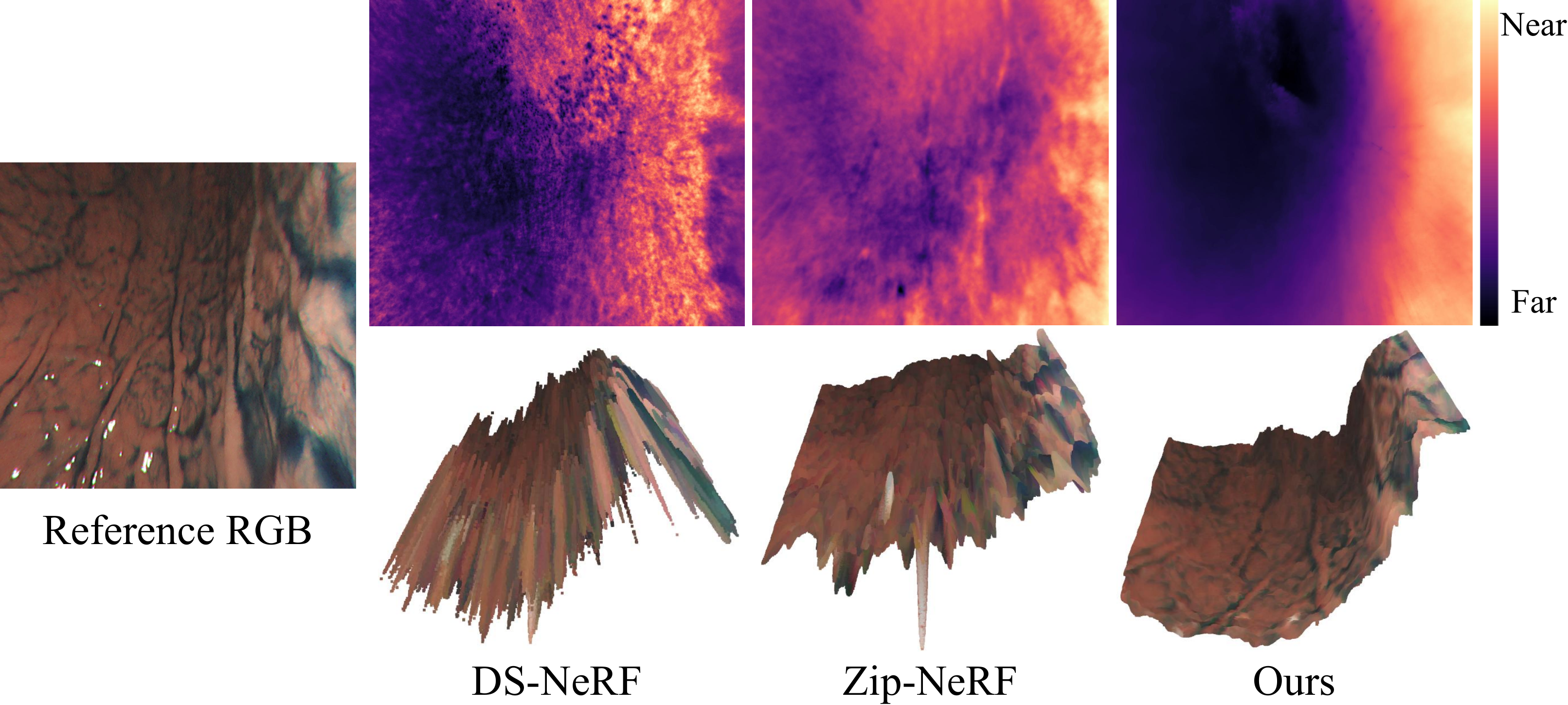}
    \caption{\textbf{The qualitative comparisons of rendered depth maps.} For each method, we present both the rendered depth map (first row) and the corresponding point cloud (second row) for better visualization.}
    \label{fig_3}
\end{figure}

We further compare the learned geometry across different methods by comparing the rendered depth maps.
Since the ground-truth depth maps are not available, only the qualitative evaluation of the learned geometry is provided as shown in Fig.~\ref{fig_3}.
For better visualization, both the rendered depth maps and their corresponding 3D point clouds are presented.
We can see that the rendered depth map of DS-NeRF exhibits significant local variations.
The depth map produced by Zip-NeRF exhibits significant errors, which shows that it is challenging for a general NeRF method solely based on a color-based loss to accurately learn geometry from monocular gastroscopic images.
Our method produces the visually best results, by incorporating geometry-based supervision on both the observed and unobserved views.

\subsection{Ablation Study}

\begin{table}[t!]
\caption{\textbf{Ablation study.} The best results are highlighted using \textbf{bold} formatting.}
\label{tab_2}
\centering
\renewcommand\arraystretch{1.2}
\resizebox{\linewidth}{!}{
\begin{tabular}{ccc|cc|cc}
\hline
\multirow{2}{*}{$L_{color}^{ob}$} & \multirow{2}{*}{$L_{depth}^{ob}$} & \multirow{2}{*}{$L_{depth}^{nv}$} & \multicolumn{2}{c|}{Seq.A}      & \multicolumn{2}{c}{Seq.B}       \\ \cline{4-7} 
                                                      &                                                       &                                                       & PSNR           & SSIM           & PSNR           & SSIM           \\ \hline
\checkmark                                                    &                                                       &                                                       & 25.07          & 0.856          & 22.78          & 0.751          \\
\checkmark                                                    & \checkmark                                                    &                                                       & 25.47          & 0.861          & 23.17          & 0.762          \\
\checkmark                                                    & \checkmark                                                    & \checkmark                                                    & \textbf{26.73} & \textbf{0.870} & \textbf{23.37} & \textbf{0.767} \\ \hline
\end{tabular}
}
\end{table}

We conducted an ablation study on the loss terms utilized in our method. The results are reported in Table~\ref{tab_2}.
Both the geometry-based losses applied to observed views and unobserved views improve the image quality of novel view synthesis, which indicates that the enhanced learned geometry consequently contributes to better image rendering results.

\section{Conclusion}
In this paper, we have employed the technique of NeRF for the task of synthesizing free-viewpoint images within the stomach.
To enhance the performance of NeRF for high-quality novel view synthesis on monocular gastroscopic data, we have augmented the original color-based training loss with our proposed geometry-based loss.
This augmentation enables effective utilization of the point cloud pre-reconstructed from the input images to constrain the implicit geometry of NeRF.
Additionally, we have proposed to apply the geometry-based supervision to randomly generated unobserved views during the training phase, which further regularizes the learned geometry and contributes to performance enhancements in novel view synthesis.
The experimental results have demonstrated that our method is capable of both high-quality novel view synthesis and geometry recovery based on monocular gastroscopy data.
One limitation of our method lies in its dependence on the precise estimation of camera poses from SfM, which is not guaranteed in the context of challenging gastroscopic data.
In the future, we plan to incorporate the refinement of camera poses into the training of NeRF to alleviate the reliance on accurately pre-estimated camera poses.
To view the video results, please visit our project page in the following link (\textcolor{blue}{\url{http://www.ok.sc.e.titech.ac.jp/res/Stomach3D/}}).

~

\noindent
\textbf{Acknowledgement}.
This work was partly supported by JSPS KAKENHI Grant Number 24K15772.




{\small
\bibliographystyle{IEEEtran}
\bibliography{egbib}

\begin{thebibliography}{10}
\providecommand{\url}[1]{#1}
\csname url@samestyle\endcsname
\providecommand{\newblock}{\relax}
\providecommand{\bibinfo}[2]{#2}
\providecommand{\BIBentrySTDinterwordspacing}{\spaceskip=0pt\relax}
\providecommand{\BIBentryALTinterwordstretchfactor}{4}
\providecommand{\BIBentryALTinterwordspacing}{\spaceskip=\fontdimen2\font plus
\BIBentryALTinterwordstretchfactor\fontdimen3\font minus \fontdimen4\font\relax}
\providecommand{\BIBforeignlanguage}[2]{{%
\expandafter\ifx\csname l@#1\endcsname\relax
\typeout{** WARNING: IEEEtran.bst: No hyphenation pattern has been}%
\typeout{** loaded for the language `#1'. Using the pattern for}%
\typeout{** the default language instead.}%
\else
\language=\csname l@#1\endcsname
\fi
#2}}
\providecommand{\BIBdecl}{\relax}
\BIBdecl

\bibitem{snavely2006photo}
N.~Snavely, S.~M. Seitz, and R.~Szeliski, ``Photo tourism: exploring photo collections in 3d,'' \emph{ACM Trans. Graphics}, vol.~25, no.~3, p. 835–846, 2006.

\bibitem{pollefeys2008detailed}
M.~Pollefeys, D.~Nist{\'e}r, J.-M. Frahm, A.~Akbarzadeh, P.~Mordohai, B.~Clipp, C.~Engels, D.~Gallup, S.-J. Kim, P.~Merrell \emph{et~al.}, ``Detailed real-time urban 3d reconstruction from video,'' \emph{IJCV}, vol.~78, pp. 143--167, 2008.

\bibitem{schonberger2016structure}
J.~L. Schonberger and J.-M. Frahm, ``Structure-from-motion revisited,'' in \emph{Proc. CVPR}, 2016.

\bibitem{mills2014hierarchical}
S.~Mills, L.~Szymanski, and R.~Johnson, ``Hierarchical structure from motion from endoscopic video,'' in \emph{Proc. of Int. Conf. on Image and Vision Computing New Zealand (IVCNZ)}, 2014.

\bibitem{sun2013surface}
D.~Sun, J.~Liu, C.~A. Linte, H.~Duan, and R.~A. Robb, ``Surface reconstruction from tracked endoscopic video using the structure from motion approach,'' in \emph{Proc. of Augmented Reality Environments for Medical Imaging and Computer-Assisted Interventions (AE-CAI)}, 2013.

\bibitem{lurie20173d}
K.~L. Lurie, R.~Angst, D.~V. Zlatev, J.~C. Liao, and A.~K.~E. Bowden, ``3d reconstruction of cystoscopy videos for comprehensive bladder records,'' \emph{Biomedical optics express}, vol.~8, no.~4, pp. 2106--2123, 2017.

\bibitem{furukawa2016shape}
R.~Furukawa, H.~Morinaga, Y.~Sanomura, S.~Tanaka, S.~Yoshida, and H.~Kawasaki, ``Shape acquisition and registration for 3d endoscope based on grid pattern projection,'' in \emph{Proc. ECCV}, 2016.

\bibitem{alcantarilla2013enhanced}
P.~F. Alcantarilla, A.~Bartoli, F.~Chadebecq, C.~Tilmant, and V.~Lepilliez, ``Enhanced imaging colonoscopy facilitates dense motion-based 3d reconstruction,'' in \emph{Proc. EMBC}, 2013.

\bibitem{widya20193d}
A.~R. Widya, Y.~Monno, K.~Imahori, M.~Okutomi, S.~Suzuki, T.~Gotoda, and K.~Miki, ``3d reconstruction of whole stomach from endoscope video using structure-from-motion,'' in \emph{Proc. EMBC}, 2019.

\bibitem{kazhdan2013screened}
M.~Kazhdan and H.~Hoppe, ``Screened poisson surface reconstruction,'' \emph{ACM Trans. Graphics}, vol.~32, no.~3, pp. 1--13, 2013.

\bibitem{mildenhall2020nerf}
B.~Mildenhall, P.~P. Srinivasan, M.~Tancik, J.~T. Barron, R.~Ramamoorthi, and R.~Ng, ``Nerf: Representing scenes as neural radiance fields for view synthesis,'' in \emph{Proc. ECCV}, 2020.

\bibitem{barron2023zipnerf}
J.~T. Barron, B.~Mildenhall, D.~Verbin, P.~P. Srinivasan, and P.~Hedman, ``Zip-nerf: Anti-aliased grid-based neural radiance fields,'' in \emph{Proc. ICCV}, 2023.

\bibitem{kajiya1984ray}
J.~T. Kajiya and B.~P. Von~Herzen, ``Ray tracing volume densities,'' \emph{ACM SIGGRAPH computer graphics}, vol.~18, no.~3, pp. 165--174, 1984.

\bibitem{wang2022neural}
Y.~Wang, Y.~Long, S.~H. Fan, and Q.~Dou, ``Neural rendering for stereo 3d reconstruction of deformable tissues in robotic surgery,'' in \emph{Proc. MICCAI}, 2022.

\bibitem{zha2023endosurf}
R.~Zha, X.~Cheng, H.~Li, M.~Harandi, and Z.~Ge, ``Endosurf: Neural surface reconstruction of deformable tissues with stereo endoscope videos,'' in \emph{Proc. MICCAI}, 2023.

\bibitem{batlle2023lightneus}
V.~M. Batlle, J.~M. Montiel, P.~Fua, and J.~D. Tard{\'o}s, ``Lightneus: Neural surface reconstruction in endoscopy using illumination decline,'' in \emph{Proc. MICCAI}, 2023.

\bibitem{reed2021dynamic}
A.~W. Reed, H.~Kim, R.~Anirudh, K.~A. Mohan, K.~Champley, J.~Kang, and S.~Jayasuriya, ``Dynamic ct reconstruction from limited views with implicit neural representations and parametric motion fields,'' in \emph{Proc. ICCV}, 2021.

\bibitem{corona2022mednerf}
A.~Corona-Figueroa, J.~Frawley, S.~Bond-Taylor, S.~Bethapudi, H.~P. Shum, and C.~G. Willcocks, ``Mednerf: Medical neural radiance fields for reconstructing 3d-aware ct-projections from a single x-ray,'' in \emph{Proc. EMBC}, 2022.

\bibitem{wu2022arbitrary}
Q.~Wu, Y.~Li, Y.~Sun, Y.~Zhou, H.~Wei, J.~Yu, and Y.~Zhang, ``An arbitrary scale super-resolution approach for 3d mr images via implicit neural representation,'' \emph{IEEE Journal of Biomedical and Health Informatics}, vol.~27, no.~2, pp. 1004--1015, 2022.

\bibitem{shen2022nerp}
L.~Shen, J.~Pauly, and L.~Xing, ``Nerp: implicit neural representation learning with prior embedding for sparsely sampled image reconstruction,'' \emph{IEEE Transactions on Neural Networks and Learning Systems}, 2022.

\bibitem{deng2022depth}
K.~Deng, A.~Liu, J.-Y. Zhu, and D.~Ramanan, ``Depth-supervised nerf: Fewer views and faster training for free,'' in \emph{Proc. CVPR}, 2022.

\end{thebibliography}
}

\end{document}